\documentclass[journal]{article}
\usepackage{graphicx} 
\usepackage{siunitx}
\usepackage{booktabs}   
\usepackage{tikz}
\usepackage{pgfplots}
\pgfplotsset{compat=1.18}
\usetikzlibrary{arrows.meta,positioning}
\title{LLM for the development of FCM}
\author{Alexis Kafantaris}
\date{June 2026}
\usepackage[margin=1in]{geometry}
\usepackage{setspace}
\usepackage{microtype} 
\usepackage{float}           
\begin{document}

\maketitle
\section*{Abstract}
This article is about the development of a fuzzy cognitive map using a local large language model. In light of recent advances, it is evident that large language models, and even local large language models, are capable of extracting quantities from textual data. In other words, a local LLM like Qwen2.5-32B, or probably larger, can accept entities as prompt input and determine relevant quantitative data as the model output. In turn, this output can be utilized for the construction of a data driven fuzzy cognitive map. Hence, this implementation is achieved, and then the model is thoroughly tested; Qwen2.5-32B is used and the data is extracted from hotel reviews from TripAdvisor. Furthermore, the extracted documents pass through the model unfiltered and then a fuzzy cognitive map is trained and evaluated. A case is made about Greek reviews where a star topology FCM is formed that indicates the preferences of the reviewers. Finally, external validation is performed to establish whether the fuzzy cognitive map can correlate the star rating of the review ---an outcome outside the model's inference scope---with its predicted satisfaction.

\section{Introduction}

\subsection{LLM}

One of the most recent advances in the scientific field is the large language model (LLM) technology. The LLM technology has redefined many industries, and the LLM integration helps automate various tasks. These tasks range from data collection, to data generation for personalized reviews~\cite{pandey2025}, to service design process planning~\cite{barandoni2026}. Another interesting use case for LLMs is the identification of topics or causal relationships. LLMs can understand context and identify topics from key words better than traditional models such as latent Dirichlet allocation~\cite{bertopic_llm_tourist2026, fuzzybertopic2026}. Hence, assisting implicitly in improving services.

An auto-regressive large language model refers to a model, Qwen2.5-32B, which uses a prediction equation as follows$$ \mathcal{L}_{CLM} = - \sum_{i=1}^{n} \log P(x_i \mid x_1, \dots, x_{i-1}; \Theta) $$

Furthermore, except for topic modeling, LLM can also identify causality; one would need a very large LLM for that, however, there is another way that is possible to create a fuzzy cognitive map (FCM). So, there are several implementations of an FCM, one explicitly from the father of FCM Bart Kosko that exploit specifically LLMs~\cite{panda2026agentic}. Kosko's idea was to automate the pipeline to generate an FCM from nouns and then connect the entities that are derived and determine causal relationships. A similar idea has also been implemented, while the LLM extracted causal relationships from various entities that were given~\cite{giabbanelli2019}.

At this point a gap is identified, that has to do with the use of the LLM for the development of an FCM. Instead of relying on the LLM to extract causality~\cite{panda2026agentic, mkhitaryan2025evolutionary}, the LLM can extract data from which an FCM can be trained. The user provides entity concepts, the LLM extracts the data, and then from the extracted elements a data driven gradient descent based FCM is formed. This might seem obvious at first, and it is a simple idea. To our knowledge, no prior work has used an LLM to extract data for the development of a data driven FCM.

\subsection{FCM}

A fuzzy cognitive map is a signed digraph; a symbolic network with concepts as nodes and fuzzy relationships as edges. By using fuzzy relations as edges it becomes interpretable, e.g. concept C1 affects concept C2 by 0.2, edge C1C2 is 0.2, or concept C3 affects concept C2 by negative 0.5, edge C3C2 is -0.5. The FCM is a shallow recurrent neural network that, instead of calculating random weights, calculates weights as system dynamics \cite{kosko1986}. In this way, one can simulate the system and, through simulation, determine the mathematical equilibrium \cite{kosko1997}. 

Fuzzy cognitive map learning methods are an interesting topic that has been thoroughly researched\cite{stach2010}. Some methods have convergence advantages, while others have interpretability. There are three main categories: Hebbian, agentic, and hybrid. Each method has advantages and disadvantages. For example, Hebbian \cite{hebb1949} based methods converge faster but are not interpretable. On the other hand, an agentic setup might lack convergence, but it is interpretable \cite{eiben2015}. Ideally, each setup is meant to be used under specific conditions. In this case a data driven fuzzy cognitive map can use quantities to train and understand patterns.

In addition, fuzzy cognitive maps are used for classification, prediction, and simulation \cite{kosko1997}. These are already key distinctions that more or less shape the architecture. There are different versions of fuzzy cognitive maps, as well as different learning methods that are shaped to address the key challenges \cite{papageorgiou2011}. Different activation functions are used depending on the objectives. The network, attempting to minimize the error, adjusts the weights, which represent causality between concepts \cite{kosko1986}.

\section{Background}\label{background}
Moreover, using fuzzy cognitive maps to model service design has excellent advantages due to the predictive capacity of FCM~\cite{tsadiras2008}. Similar attempts have been made regarding both entity extraction and sentiment analysis. In fact, sentiment analysis is probably one of the most researched topics, which helps with the implementation of the FCM. For example, fsQCA~\cite{fsqca_cuba2024} and the ABSA fuzzy mapping~\cite{bi2019type2, liu2017ranking} are very close to that idea conceptually. Instead of determining concept entities, the fsQCA determines all possible configurations of given concepts~\cite{qin2025}. Here, entity extraction has again been achieved with an LLM or with traditional sentiment analysis models. Moreover, a similar pattern has been researched for ABSA, where sentiment is extracted using an LLM or traditional models and then hotel reviews are analyzed~\cite{fuzzy_absa_lstm2025, althubiti2025}.

Another attempt has been made using evolutionary algorithms to create a fuzzy cognitive map~\cite{mkhitaryan2025evolutionary}. To achieve that, a  six step pipeline is implemented and then an agentic FCM is created. Implementing the CMA-ES algorithm on causal relationships that the models provide-validate, and then evolving the correct edges based on three artificial datasets works. However, one major drawback is that this model relies on expensive LLMs to determine causality. In addition, using the data to determine whether an FCM creation can be achieved in a more efficient way has not been explored. Eventually, there are some ways to automate the creation of fuzzy cognitive maps. It is believed that a better idea is to set entities and extract data for a gradient descent to find causality.

To achieve data structuring, aspect-based sentiment analysis (ABSA) is performed on a fixed set of aspects, assigning a continuous sentiment score to each aspect through zero-shot LLM prediction. In particular, aspect category sentiment analysis (ACSA). This is due to the fact that the LLM annotation has already been proven and established by GPT as there has been a good prior~\cite{gilardi2023chatgpt}; from GPT a step forward was to show that local LLMS, Qwen2.5 too, can perform ACSA~\cite{multiling_absa_llm2025}. As such, Qwen2.5-32B was used instead of GPT and is perceived as a valid alternative.  However, the results might not be SOTA accurate due to various constraints. Still, the adoption of local LLM seems reasonable as it achieves high accuracy~\cite{absa_slr2024}. The aspect set itself was grounded in Hontology, a multilingual accommodation-sector ontology, from which the top-level review-relevant categories were selected~\cite{chaves2012hontology}.

\section{Methodology}

\subsection{Data Extraction}
Data were sourced from TripAdvisor. There were 1505 Greek hotel reviews in total, for more than 100 hotels. Then from review data, a prompt was assigned to derive tabular like quantities from texts. 

Here, a few things are worth noting, things that have to do with the choices and the design process. Firstly, several models were tested while trying to figure out the smallest capable model. Having stated that, one started testing from Qwen2.5-0.5B to Mistral-7B before reaching Qwen2.5-14B and, for the final concept set, Qwen2.5-32B. Smaller models could not extract correctly the entities. It was either too much or too little extraction, with too much being a noisy signal that plateaued on a given R2 and then it practically failed to explain the phenomena. The 'dirty' signal could only account for more dirty signal until it was noticed; the primary sign was that the sentence annotations did not make sense and that the R2 test was a plain flat line and did not have an upward trend.

Secondly, the tests were conducted at both sentence level and at the full review level. Here, the results were pretty similar, similar enough to suggest that both schemes work. However, an interesting detail is that per review level, the R2 test resulted in a higher score. Hence, it was determined that due to the higher R2 and the fewer preprocessing requirements per review level, sentiment analysis is better per review level. In the other case, sentences that carried a single entity had to be filtered because they did not contribute to the learning of the FCM. Fortunately, the same model could still impute the sentences, which makes the sentiment extraction a case of model parameter size. That is, larger models can extract entities better than smaller models for which after a point the entities can be used for the creation of an FCM.

Finally, the concept set itself was expanded and re-extracted with the larger Qwen2.5-32B model, grounded in the top-level categories of Hontology (Section~\ref{background}). Initially, there were seven handpicked categories based on experience about hotel reviews. For these Qwen2.5-14B was sufficient; later, a decision was made to use an ontology  model about tourism. There the Qwen2.5-14B could not successfully entangle the concepts, hence, Qwen2.5-32B was used for nine concepts. 

\subsection{FCM creation}
The relevant entities for which data was supposed to be derived were assigned, and the data was extracted. Then a data driven FCM was created; the FCM Equations are as follows:
\[
H_{t+1} = \tanh(H_t W),
\]
The driver$\rightarrow$satisfaction weights are learned using gradient descent. The weights move against the gradient of the squared prediction
error by a learning rate $\eta$:
\[
W \;\leftarrow\; W \;-\; \eta\,\nabla_{W}\,\mathcal{L},
\qquad
\mathcal{L} \;=\; 1/2\,||H_T - H_{\text{target}}||^{2}.
\]

The gradient descent algorithm is a mathematical tool for finding the fit of data in a model. Using the derivative of the error, the program descends in the best direction for error minimization. In other words, by moving towards the direction of the gradient, the program reduces the error of the target by a step eta; as the error decreases, the model is closer to the target state. Due to the mathematical background of this method, it is a very efficient program. However, this method has some issues; to rely on a step each time means that the algorithm can get stuck at a local minimum that is deeper than a step, that is, having the error larger than the step. On the other hand, with too large a step, the algorithm gets unstable and does not converge either. 

Here is the full pipeline; it consists of five steps.
\newline
\begin{tikzpicture}[
node distance=0.9cm,
box/.style={
draw,
rounded corners,
minimum width=2.0cm,
minimum height=0.9cm,
align=center,
font=\small,
fill=blue!8
},
>=Stealth
]

\node[box] (reviews) {TripAdvisor\\Reviews};
\node[box,right=of reviews] (acsa) {ACSA};
\node[box,right=of acsa] (table) {Feature\\Table};
\node[box,right=of table] (train) {FCM\\Training};
\node[box,right=of train] (map) {Learned\\FCM};

\draw[->,thick] (reviews)--(acsa);
\draw[->,thick] (acsa)--(table);
\draw[->,thick] (table)--(train);
\draw[->,thick] (train)--(map);

\end{tikzpicture}
\section{Results}

Given that the LLM has an important role in the development of data driven FCM, a series of tests was conducted. The first test was a held-out $R^2$ test, checking whether the model generalized. The second was an ablation study on concepts. The third was a null permutation, checking that the $R^2$ score is not achieved by chance. The fourth was a data stream to determine whether more data achieve better results. The fifth was a comparison with the regression baseline and a mean estimator. The sixth was a $k$-fold validation to determine whether the system is stable. Table~\ref{tab:concepts} summarizes the corpus, and Table~\ref{tab:battery} collects the headline numbers for the entire battery; the subsections below break each test out individually.

\begin{table}[ht]\centering
\caption{Concept frequency in the Greek-reviewer corpus (1505 reviews, 1491 after filtering to rows with a satisfaction rating and at least one driver). One row per review; mean shown for present (non-zero) cells only.}
\label{tab:concepts}\begin{tabular}{lrr}\toprule
Concept & Present (rows) & Mean sentiment \\ \midrule
Cleanliness & 554 & +0.63 \\
Staff & 1190 & +0.66 \\
Location & 988 & +0.74 \\
Breakfast & 688 & +0.52 \\
Noise & 177 & -0.14 \\
Value & 373 & +0.10 \\
Comfort & 975 & +0.52 \\
Amenities & 733 & +0.52 \\
Check-in & 182 & +0.27 \\
Satisfaction & 1500 & +0.64 \\
\bottomrule\end{tabular}\end{table}

\begin{table}[ht]\centering
\caption{Evaluation battery for the Greek-reviewer FCM (per-review; 70/30 held-out; permutation null over 200 target shuffles).}
\label{tab:battery}\begin{tabular}{lc}\toprule
Test & Value \\ \midrule
Held-out $R^2$ (FCM) & $+0.795$ \\
Held-out $R^2$ (linear regression) & $+0.669$ \\
Held-out $R^2$ (mean predictor) & $-0.000$ \\
$k$-fold $R^2$ (5 folds) & $+0.782 \pm 0.022$ \\
Permutation null mean $R^2$ & $-0.324$\\
Permutation $p$-value & $0.0050$ \\
\bottomrule\end{tabular}\end{table}
\subsection{R2 test}
The model's ability to generalize is assessed on a 70/30 held-out split. The model learns how 70 percent of the data connects, and then it tries to explain the remaining 30 percent. More than 0.6 R2 is considered to have high predictive capacity.

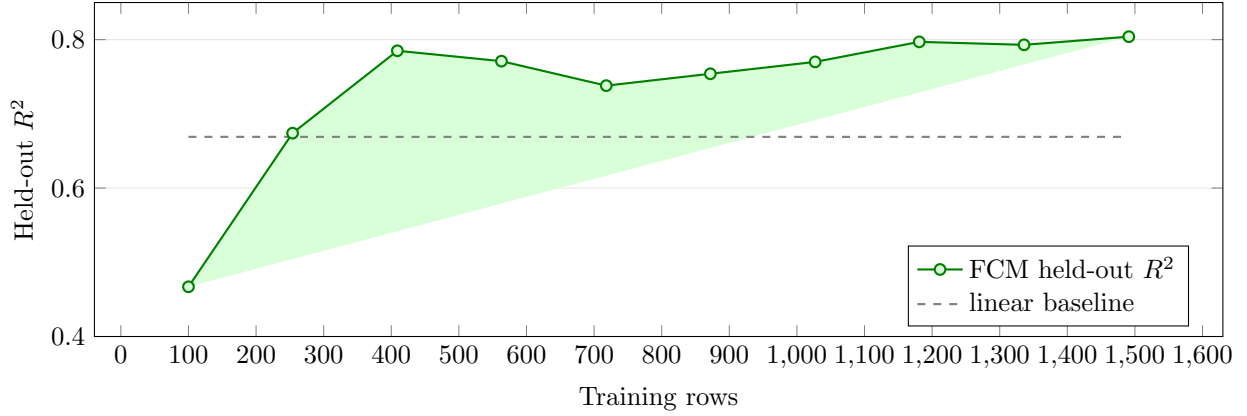
\begin{figure}[ht]
\centering
\begin{tikzpicture}
\begin{axis}[
    width=\linewidth, height=6cm,
    ymin=0.4, ymax=0.85,
    xlabel={Training rows}, ylabel={Held-out $R^2$},
    ymajorgrids, grid style={gray!20},
    legend pos=south east, legend cell align=left,
]
\addplot[mark=*,thick,green!50!black,fill=green!15] coordinates {
    (100,0.467) (254,0.674) (409,0.785) (563,0.771) (718,0.738)
    (872,0.754) (1027,0.770) (1181,0.797) (1336,0.793) (1491,0.804)
};
\addlegendentry{FCM held-out $R^2$}
\addplot[dashed,thick,gray] coordinates {(100,0.669) (1491,0.669)};
\addlegendentry{linear baseline}
\end{axis}
\end{tikzpicture}
\caption{Held-out $R^2$ versus training rows for the nine-driver model. After an unstable start on very few rows, performance settles around $0.75$--$0.80$ and remains above the linear regression baseline ($0.67$); the curve is noisy in the mid-range and trends upward as more reviews are fed to it.}
\label{fig:r2vsdata}
\end{figure}

\subsection{Ablation study}
To determine whether the entities are important or not, an ablation study was performed. Each driver was removed in turn and the held-out $R^2$ loss recorded; then each component was ranked based on the impact that it has on the R2 score of the model.

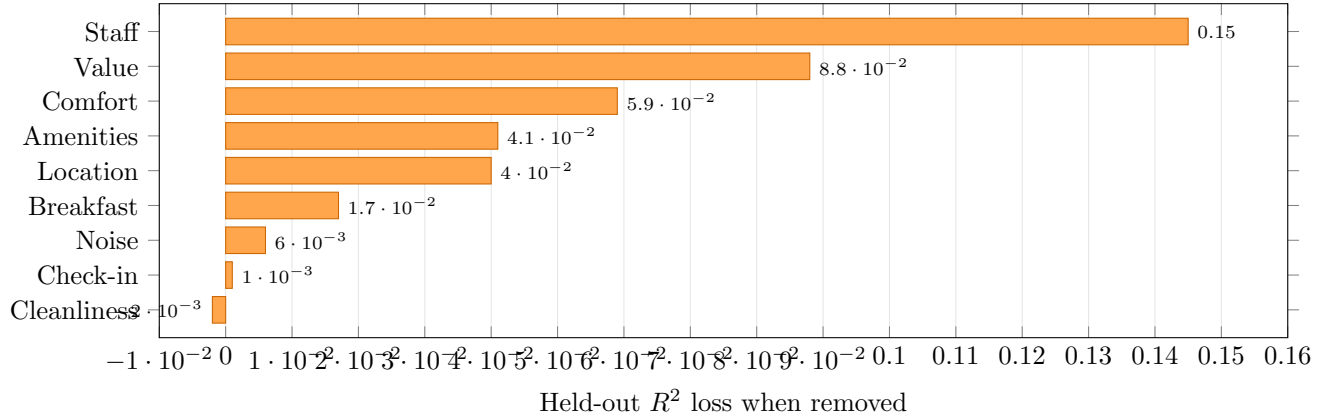
\begin{figure}[ht]
\centering
\begin{tikzpicture}
\begin{axis}[
    width=\linewidth, height=6cm,
    xbar, bar width=10pt,
    xmin=-0.01, xmax=0.16,
    xlabel={Held-out $R^2$ loss when removed},
    symbolic y coords={Cleanliness,Check-in,Noise,Breakfast,Location,Amenities,Comfort,Value,Staff},
    ytick=data,
    nodes near coords, every node near coord/.append style={font=\scriptsize},
    xmajorgrids, grid style={gray!20},
]
\addplot[fill=orange!70,draw=orange!80!black] coordinates {
    (0.145,Staff) (0.088,Value) (0.059,Comfort) (0.041,Amenities)
    (0.040,Location) (0.017,Breakfast) (0.006,Noise) (0.001,Check-in) (-0.002,Cleanliness)
};
\end{axis}
\end{tikzpicture}
\caption{Ablation: staff is the most important parameter for explaining satisfaction, followed by value and comfort. Check-in and cleanliness contribute negligibly once the other drivers are present.}
\label{fig:ablation}
\end{figure}
\newpage
\subsection{Null Permutation}
To confirm the fit is not an artifact, the satisfaction target was shuffled over 200 permutations, and the FCM re-fit each time. Essentially, the model tried to explain the data in random configurations. If the results are random, then when one shuffles the target, the variance is still explained, as it was chance. On the other hand, if the shuffled data is not explained, it is not chance, as it was shuffled and was not explained. Now, whether this poses a statistically significant change remained unanswered, and indeed it is statistically significant. Or, to rephrase it, the predictive capability comes from the data.

\subsection{Data stream}
Held-out $R^2$ was tracked to test whether additional data improves the fit. After an unstable start on very few rows, performance settles around $0.75$--$0.80$ at the full 1491 rows and stays above the linear baseline throughout; (Figure~\ref{fig:r2vsdata}). Here, a trend was searched, or whether more data increase the R2 score, which it does. Calculating the R2 while feeding data indicated that for the first data a spike occurs and then for the other there is an upward trend nonetheless.

\subsection{Baseline Comparison}
The FCM was bench-marked against a linear-regression baseline and a mean estimator on the same held-out split. The purpose of this test was to determine whether an FCM is better than linear regression or a mean estimator to understand the reviews.

In addition, a second and more thorough comparison was devised; the model was now compared with other Python-based baselines. 
\begin{table}[ht]\centering
\caption{FCM comparison to Python baselines on the same LLM-extracted aspect features (per-review, 5-fold $R^2$).}
\label{tab:model_comparison}\begin{tabular}{lcc}\toprule
Model & $k$-fold $R^2$ & Interpretable causal map \\ \midrule
XGBoost              & $+0.856 \pm 0.025$ & No \\
Random Forest        & $+0.850 \pm 0.032$ & No \\
MLP                  & $+0.827 \pm 0.020$ & No \\
\textbf{FCM (ours)}  & $+0.782 \pm 0.022$ & \textbf{Yes} \\
Linear regression    & $+0.667 \pm 0.026$ & Partial \\
Mean estimator       & $-0.005 \pm 0.005$ & --- \\
\bottomrule\end{tabular}\end{table}
That being said, the FCM does not beat the MLP, XGBoost, or Random Forest models, but it has other advantages. The point of traditional machine learning is raw predictive power; in this instance, the interpretability is required. Furthermore, the signal of the data labels is real and now other programs like XGBoost point at it. Indeed, the FCM is not the best predictor, but it still provides a good result. It also yields clarity about the choices it has made; the point of the FCM was to personalize experiences and understand user feedback.

\subsection{K fold validation}
The stability of the system was checked with a 5-fold cross-validation. Data were shuffled, and then the FCM weights were determined. Using parts of the data, different weight estimates were expected, which makes it interesting. Assuming that the system is stable, the standard deviation of the weights should be more or less low; having a high standard deviation suggests that the system is unstable and that new data might corrupt the weights. However, this was not the case, the system was stable and sound.

\begin{table}[ht]\centering
\caption{Learned FCM driver$\rightarrow$satisfaction weights (per-review; $k$-fold mean $\pm$ s.d.\ over 5 folds), bounded to $[-1,1]$.}
\label{tab:weights}\begin{tabular}{lc}\toprule
Driver & Weight (mean $\pm$ s.d.) \\ \midrule
Staff & $+0.649 \pm 0.013$ \\
Value & $+0.558 \pm 0.020$ \\
Comfort & $+0.505 \pm 0.012$ \\
Amenities & $+0.416 \pm 0.015$ \\
Location & $+0.357 \pm 0.013$ \\
Breakfast & $+0.298 \pm 0.011$ \\
Noise & $+0.225 \pm 0.011$ \\
Cleanliness & $+0.169 \pm 0.013$ \\
Check-in & $+0.130 \pm 0.024$ \\
\bottomrule\end{tabular}\end{table}

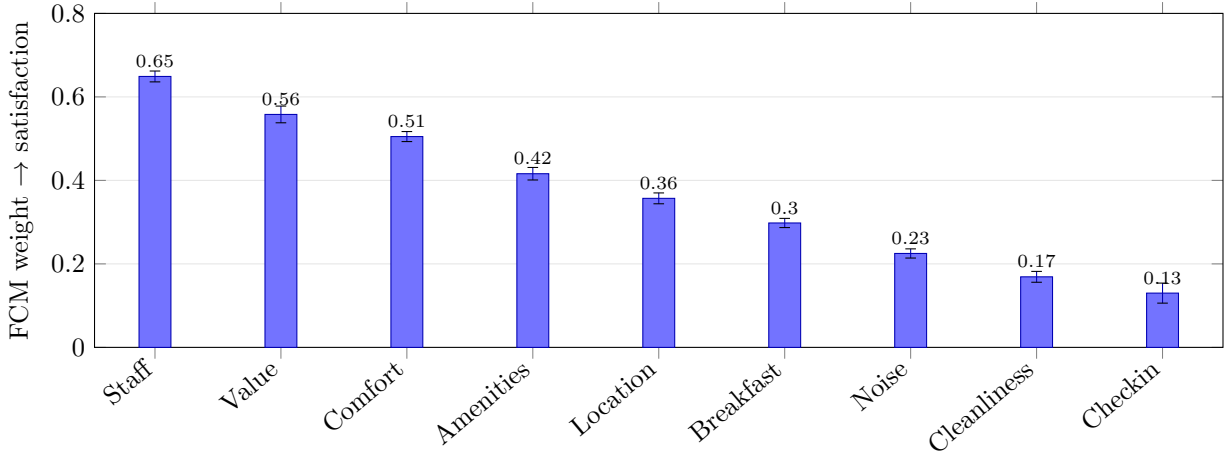
\begin{figure}[ht]
\centering
\begin{tikzpicture}
\begin{axis}[
    width=\linewidth, height=6cm,
    ybar, bar width=12pt,
    ymin=0, ymax=0.80,
    ylabel={FCM weight $\rightarrow$ satisfaction},
    symbolic x coords={Staff,Value,Comfort,Amenities,Location,Breakfast,Noise,Cleanliness,Checkin},
    xtick=data, x tick label style={rotate=40,anchor=east},
    enlarge x limits=0.06,
    error bars/y dir=both, error bars/y explicit,
    nodes near coords, every node near coord/.append style={font=\scriptsize},
    ymajorgrids, grid style={gray!20},
]
\addplot[fill=blue!55,draw=blue!70!black] coordinates {
    (Staff,0.649)      +- (0,0.013)
    (Value,0.558)      +- (0,0.020)
    (Comfort,0.505)    +- (0,0.012)
    (Amenities,0.416)  +- (0,0.015)
    (Location,0.357)   +- (0,0.013)
    (Breakfast,0.298)  +- (0,0.011)
    (Noise,0.225)      +- (0,0.011)
    (Cleanliness,0.169)+- (0,0.013)
    (Checkin,0.130)    +- (0,0.024)
};
\end{axis}
\end{tikzpicture}
\caption{Learned FCM driver$\rightarrow$satisfaction weights (per-review, $k$-fold mean $\pm$ s.d.), nine-driver model.}
\label{fig:weights}
\end{figure}

\newpage
\section{Case study}
\subsection{External validation of Greek hotel reviews}
Here is an FCM model for Greek reviewers according to TripAdvisor; one can see that satisfaction is the main driver in Greek hotel reviews while the primary attributes for satisfaction are staff and value. Moreover, the FCM has a star topology meaning there is a central node connected to other concept nodes.
\begin{figure}[ht]
\centering
\begin{tikzpicture}[
    >={Stealth[length=2.2mm]},
    node distance=0mm,
    concept/.style={draw, rounded corners=2pt, minimum width=17mm,
                    minimum height=6.5mm, font=\tiny, align=center,
                    fill=blue!8, draw=blue!55!black},
    target/.style={draw, circle, minimum size=19mm, font=\small\bfseries,
                   align=center, fill=teal!18, draw=teal!55!black},
    wlab/.style={font=\tiny, fill=white, inner sep=1pt},
]
\node[target] (sat) at (0,0) {satis-\\faction};

\node[concept] (staff) at ( 0.00, 4.40) {staff};
\node[concept] (val)   at ( 2.83, 3.37) {value};
\node[concept] (com)   at ( 4.33, 0.76) {comfort};
\node[concept] (am)    at ( 3.81,-2.20) {amenities};
\node[concept] (loc)   at ( 1.50,-4.13) {location};
\node[concept] (brk)   at (-1.50,-4.13) {breakfast};
\node[concept] (noi)   at (-3.81,-2.20) {noise};
\node[concept] (cln)   at (-4.33, 0.76) {cleanliness};
\node[concept] (chk)   at (-2.83, 3.37) {check-in};

\draw[->, line width=2.0pt,  blue!70!black] (staff) -- (sat) node[wlab,pos=0.62]{0.65};
\draw[->, line width=1.79pt, blue!66!black] (val)   -- (sat) node[wlab,pos=0.62]{0.56};
\draw[->, line width=1.67pt, blue!62!black] (com)   -- (sat) node[wlab,pos=0.62]{0.51};
\draw[->, line width=1.46pt, blue!57!black] (am)    -- (sat) node[wlab,pos=0.62]{0.42};
\draw[->, line width=1.32pt, blue!52!black] (loc)   -- (sat) node[wlab,pos=0.62]{0.36};
\draw[->, line width=1.19pt, blue!48!black] (brk)   -- (sat) node[wlab,pos=0.62]{0.30};
\draw[->, line width=1.02pt, gray!80!black] (noi)   -- (sat) node[wlab,pos=0.62]{0.23};
\draw[->, line width=0.89pt, gray!70!black] (cln)   -- (sat) node[wlab,pos=0.62]{0.17};
\draw[->, line width=0.80pt, gray!60!black] (chk)   -- (sat) node[wlab,pos=0.62]{0.13};
\end{tikzpicture}

\caption{Data-driven fuzzy cognitive map for Greek-reviewer hotel satisfaction, nine-driver ontology-grounded model.}
\label{fig:fcm}
\end{figure}
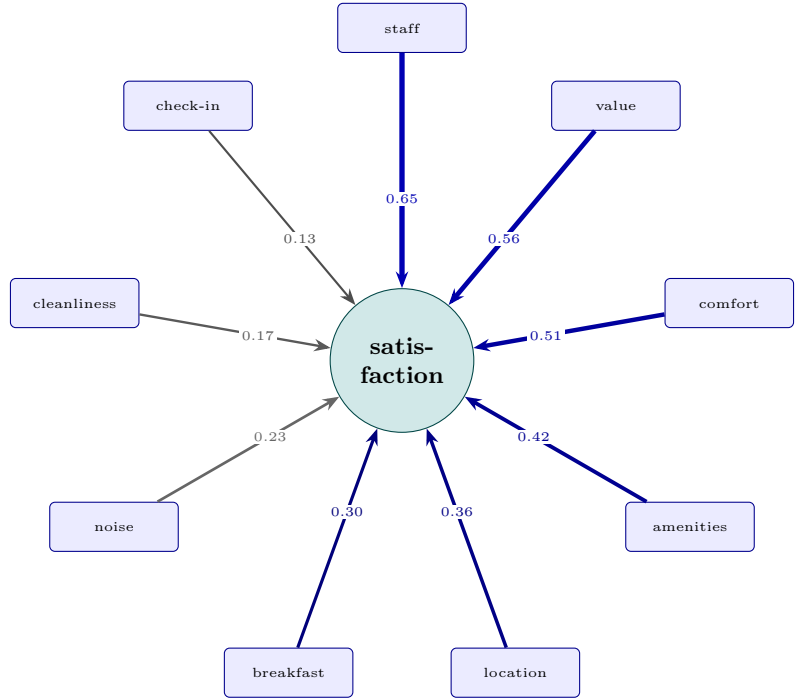\newpage

The held-out battery established that the FCM generalizes. The model is able to predict trends in the data. In addition, an external validation is performed to determine whether that predictive capability reflects a real signal. External validation takes place using the star ratings of the reviews. Neither the LLM nor the FCM has at any point come across star ratings, only passages and entities. However, it is indicated that the higher the satisfaction explicitly or implicitly through other values, the higher the overall star rating of the hotel review.

After examining the data, a correlation between the stars of the reviews and the model satisfaction is hypothesized. Satisfaction as the main driver of the star FCM topology is affected by many factors and works as the primary influence of the review according to the model. The correlation between reviews and satisfaction found was assessed using the Pearson correlation coefficient. Moreover, there exists a correlation and the more reviews per cluster are, the higher the coefficient gets. Something that might seem counterintuitive is that as the hotels become fewer they seem to provide the best signal. Considering that the highest correlation coefficient is 0.81 it is evident that the signal is real.

\begin{figure}[ht]
\centering
\includegraphics[width=0.8\linewidth]{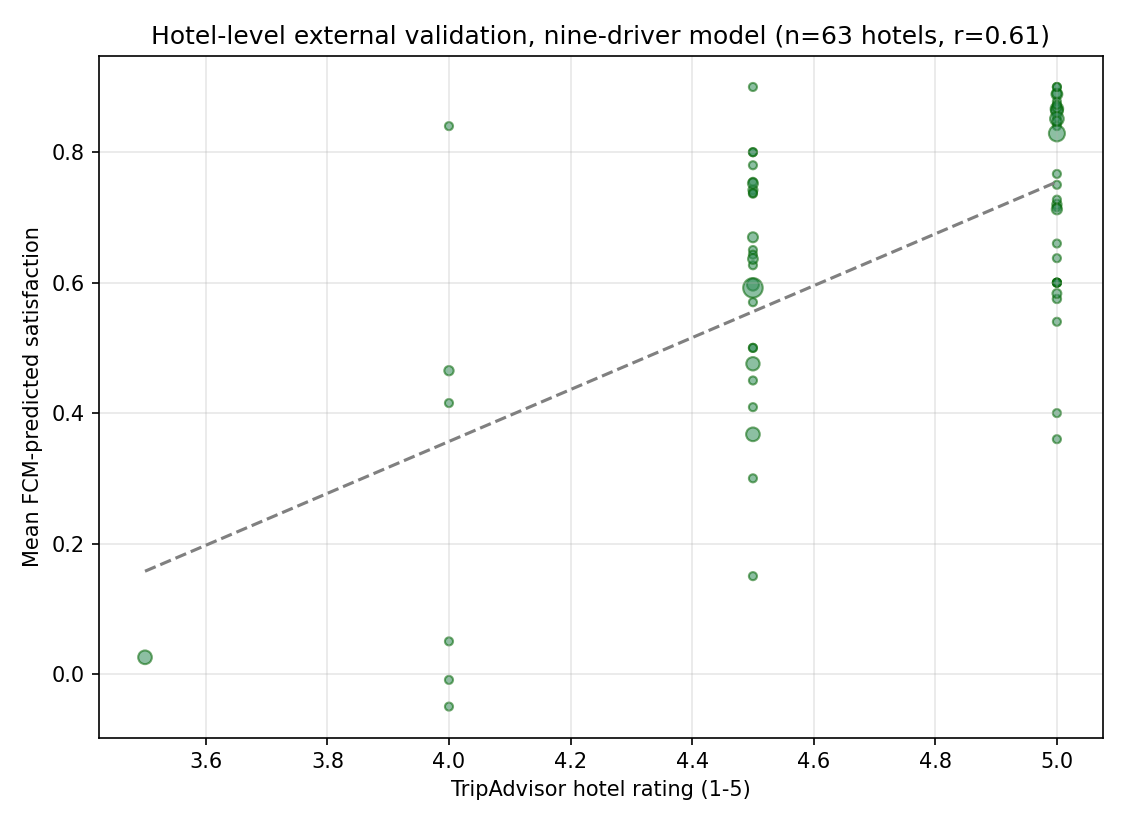}
\caption{External validation, nine-driver model. Each data point represents a hotel cluster ($n=63$, $\geq 5$ Greek reviews each), and the Pearson correlation coefficient is used to determine whether the predicted satisfaction results in higher star ratings.}
\label{fig:external}
\end{figure}

\begin{table}[ht]\centering
\caption{Hotel-level external validation (nine-driver model): correlation between mean FCM-predicted satisfaction and the official TripAdvisor hotel rating, at three minimum per-hotel review counts. As per-hotel estimates stabilize, agreement strengthens.}
\label{tab:external}\begin{tabular}{lccc}\toprule
Min.\ reviews / hotel & Hotels ($n$) & Pearson $r$ ($p$) & Spearman $\rho$ ($p$) \\ \midrule
$\geq 1$  & 107 & $+0.328\ (0.0006)$ & $+0.354\ (0.0002)$ \\
$\geq 5$  & 63  & $+0.610\ (<0.001)$ & $+0.481\ (<0.001)$ \\
$\geq 10$ & 38  & $+0.807\ (<0.001)$ & $+0.703\ (<0.001)$ \\
\bottomrule\end{tabular}\end{table}

\newpage
\section{Discussion}

Some things are worth further discussion of the results; an interesting observation was made during the ablation study. First, almost all the reviews were valid. From 1505 mined texts, 1500 carried a satisfaction rating and 1491 also carried at least one driver, which is close to a hundred percent. In this way, the causal map of the Greek-reviewer is stated to reflect the collective of all reviews, thus providing a more accurate model. Furthermore, the signal was highest for staff and value, both by edge magnitude and by ablation loss. To clarify this, it is important to understand ablation loss vs. edge magnitude. High edge magnitude does not necessarily mean high ablation loss, but high ablation loss means that the driver is more irreplaceable for the model, that is, removing it degrades the model more.

Another interesting story is the LLM, because other models were tried for ACSA but eventually were not used. Basically, it was necessary to determine that the labels are valid. Although this model is validated empirically contemporary research suggests that such practice takes place and results in accurate labels~\cite{multiling_absa_llm2025}. Had the labels not been valid, then the model would still pass the tests and there would be no way to spot it apart from reviewing it manually. That is the way it was initially attempted; at the start, Qwen2.5-0.5B was used and the results were good. Upon closer inspection, however, the model was deemed wrong, as it consistently hallucinated and also mislabeled some aspects.

Then Mistral-7B was used, which also somewhat hallucinated. No more than 20 examples were needed to determine whether the model hallucinates or mislabels the aspect dimensions. So, a different prompt was used, but still the gains were insignificant. Eventually, a larger model could perform the ACSA better, as it stands factually that larger models tend to perform better at NLP tasks~\cite{multiling_absa_llm2025}. And so Qwen2.5-32B was used for the initial seven-concept aspect-based sentiment analysis; here it is not certain whether it was the size or the different model that fixed the problem, it might be both. Nonetheless, a hundred of the reviews were annotated by Claude too, and there was ninety percent model agreement and ninety-eight percent sign or sentiment agreement.

At this point it is noted that Valence Aware Dictionary and Reasoner (VADER) was also used for the same task. It seemed like a good idea to extract the sentiment and use it to train an FCM so the first attempt was done with VADER. However, the result was a failure and then quickly it became apparent that the issues were the limitations of the VADER framework. VADER could only extract one category and failed during mixed sentiment extraction. Furthermore, it was a cheap method that might work in a different framework, other than hotel reviews.

Which brings the discussion to one of the actual limitations; the use of a local large language model is both expensive and limiting. Having asserted that, the task can either be performed by a local LLM or by a SOTA LLM like Claude Fable. And that is a bit problematic. Even having half of the model parameters is still an expensive task that requires a GPU and will only run on a limited set of reviews, mostly hundreds of thousands. After the hundred-thousand point of reviews, it requires hours, and so it should better be left off. Now, assuming one wants to use a SOTA LLM, the price starts to rise.

Lastly, another concern is the entities that the model can extract, assuming it is a local model. Both the quality, i.e. what entity dimensions the ACSA covers, and the quantity, i.e. how many entities can be extracted. That is important because more entities mean higher resolution and better recommendations.
\section{Conclusion}
Finally, in this paper ACSA is used to score predefined entities from text and then create tabular data. From the tabular data, a data-driven fuzzy cognitive map is trained using gradient descent; having satisfaction as the main driver and several key dimensions such as staff and value as primary drivers, an FCM that describes the experience is created. Then validation takes place for the target, and a case study is performed to assess the main driver, i.e. satisfaction. The main driver is compared with star reviews to validate the model, which correlates sufficiently to suggest that the extraction pipeline works. Future research should investigate more entities and try to use the FCM as a personalized recommendation tool.

\newpage
\section*{Acknowledgment}
It is acknowledged that this paper is part of a PhD dissertation, fuzzy optimization of information transmission in service design process currently done in (Athens University of Economics and Business) AUEB. It is also written with the help of Dr. Dimitris Kardaras who was the supervisor of the specific subject in AUEB and also interested in service optimization.


\newpage
\begin{thebibliography}{99}


\bibitem{pandey2025}
Pandey, P. \& Singh, J.P. (2025).
Generating product reviews from aspect-based ratings using large language models.
\textit{Journal of Retailing and Consumer Services}, 84, 104244.

\bibitem{barandoni2026}
Barandoni, S., Chiarello, F., Cascone, L., Marrale, E. \& Puccio, S. (2026).
Automating customer needs analysis: A comparative study of large language models in the travel industry.
\textit{Technological Forecasting and Social Change} (in press).

\bibitem{qin2025}
Qin, Y., Luo, C. \& Ngai, E.W.T. (2025).
Deconstructing customer satisfaction recipes: A dynamic configurational framework leveraging the power of online reviews in tourism contexts.
\textit{Tourism Management}, 110, 105181.

\bibitem{althubiti2025}
Althubiti, K., Alhamadani, A., Khan, M. \& Shah, M.G.H. (2025).
Unveiling negative memorable experiences of hotel guests: An innovative algorithmic analysis.
\textit{International Journal of Hospitality Management}, 126, 104087.

\bibitem{bertopic_llm_tourist2026}
(2026). Identifying tourist preferences from online reviews: An integrated BERTopic--LLM approach with bidirectional validation.
\textit{Journal of Open Innovation: Technology, Market, and Complexity}.

\bibitem{fuzzybertopic2026}
(2026). Fuzzy BERTopic: A neural multi-topic modeling approach based on BERT and fuzzy clustering.
\textit{Knowledge-Based Systems}.

\bibitem{giabbanelli2019}
Pillutla, V.S. \& Giabbanelli, P.J. (2019).
Iterative generation of insight from text collections through mutually reinforcing visualizations and fuzzy cognitive maps.
\textit{Applied Soft Computing}, 76, 459--478.

\bibitem{kosko1986}
B. Kosko,
``Fuzzy cognitive maps,''
\textit{International Journal of Man-Machine Studies}, vol.~24, no.~1, pp.~65--75, 1986.

\bibitem{kosko1997}
B. Kosko,
\textit{Fuzzy Engineering},
Prentice Hall, 1997.

\bibitem{stach2010}
W. Stach, L. Kurgan, and W. Pedrycz,
``Expert-based and computational methods for developing fuzzy cognitive maps,''
in \textit{Fuzzy Cognitive Maps: Advances in Theory, Methodologies, Tools and Applications},
M. Glykas, Ed., Studies in Fuzziness and Soft Computing, vol.~247,
Springer, Berlin, Heidelberg, 2010, pp.~23--41.

\bibitem{papageorgiou2011}
E. I. Papageorgiou and P. P. Groumpos,
``A new hybrid method using evolutionary algorithms to train fuzzy cognitive maps,''
\textit{Applied Soft Computing}, vol.~5, no.~4, pp.~409--431, 2005.

\bibitem{tsadiras2008}
A. K. Tsadiras,
``Comparing the inference capabilities of binary, trivalent and sigmoid fuzzy cognitive maps,''
\textit{Information Sciences}, vol.~178, no.~20, pp.~3880--3894, 2008.

\bibitem{hebb1949}
D. O. Hebb,
\textit{The Organization of Behavior: A Neuropsychological Theory},
Wiley, New York, 1949.

\bibitem{eiben2015}
A. E. Eiben and J. E. Smith,
\textit{Introduction to Evolutionary Computing},
2nd ed., Springer, 2015.

\bibitem{panda2026agentic}
A.~K. Panda, O.~Adigun, and B.~Kosko,
``The Agentic Leash: Extracting Causal Feedback Fuzzy Cognitive Maps with LLMs,''
arXiv preprint arXiv:2601.00097, 2026.

\bibitem{bi2019type2}
J.-W. Bi, Y. Liu, and Z.-P. Fan,
``Representing sentiment analysis results of online reviews using interval type-2 fuzzy numbers and its application to product ranking,''
\textit{Information Sciences}, vol.~504, pp.~293--307, 2019.

\bibitem{mkhitaryan2025evolutionary}
Schuerkamp, R. \& Giabbanelli, P.J. (2025).
Guiding evolutionary algorithms with large language models to learn fuzzy cognitive maps.
\textit{Neural Computing and Applications}, 37, 11891--11908.

\bibitem{liu2017ranking}
Y. Liu, J.-W. Bi, and Z.-P. Fan,
``A method for ranking products through online reviews based on sentiment classification and interval-valued intuitionistic fuzzy TOPSIS,''
\textit{International Journal of Information Technology \& Decision Making}, vol.~16, no.~6, pp.~1497--1522, 2017.

\bibitem{fsqca_cuba2024}
V. Perdomo-Verdecia, P. Garrido-Vega, and M. Sacrist\'{a}n-D\'{i}az,
``An fsQCA analysis of service quality for hotel customer satisfaction,''
\textit{International Journal of Hospitality Management}, vol.~122, 103793, 2024.

\bibitem{fuzzy_absa_lstm2025}
M. Sivakumar and S. R. Uyyala,
``Aspect-based sentiment analysis of mobile phone reviews using LSTM and fuzzy logic,''
\textit{International Journal of Data Science and Analytics}, 2021.

\bibitem{gilardi2023chatgpt}
F. Gilardi, M. Alizadeh, and M. Kubli,
``ChatGPT outperforms crowd workers for text-annotation tasks,''
\textit{Proceedings of the National Academy of Sciences}, vol.~120, no.~30, e2305016120, 2023.

\bibitem{multiling_absa_llm2025}
C. Wu, B. Ma, Z. Zhang, N. Deng, Y. He, and Y. Xue,
``Evaluating zero-shot multilingual aspect-based sentiment analysis with large language models,''
\textit{International Journal of Machine Learning and Cybernetics}, 2025.

\bibitem{absa_slr2024}
Y. C. Hua, P. Denny, J. Wicker, and K. Taskova,
``A systematic review of aspect-based sentiment analysis: domains, methods, and trends,''
\textit{Artificial Intelligence Review}, vol.~57, no.~11, 296, 2024.

\bibitem{chaves2012hontology}
M.~S. Chaves, L. Freitas, and R. Vieira,
``Hontology: a multilingual ontology for the accommodation sector in the tourism industry,''
in \textit{Proc. 4th Int. Conf. on Knowledge Engineering and Ontology Development (KEOD)}, 2012.














\end{thebibliography}
\end{document}